\documentclass[conference,letterpaper]{IEEEtran}

\IEEEoverridecommandlockouts

\usepackage{amsmath,amssymb,amsfonts}
\usepackage[left=1.91cm, right=1.91cm, top=2.54cm, bottom=1.91cm]{geometry}

\usepackage{graphicx}
\usepackage{tikz}
\usepackage{pgfplots}
\pgfplotsset{compat=newest}
\DeclareGraphicsExtensions{.pdf,.png}
\usepackage{epstopdf}

\usepackage{algorithm}
\usepackage{algorithmic}

\usepackage{subcaption}

\usepackage{float}
\usepackage{csvsimple}
\usepackage{pgfplotstable}
\usepackage{lscape}

\usepackage{colortbl}
\usepackage{url}

\usepackage[style=ieee, backend=biber]{biblatex}
\usepackage{xcolor}

\definecolor{customBlue}{RGB}{164,203,232}
\definecolor{customGreen}{RGB}{164,226,192}

\begin{document}

\title{Vector Map Quality Metrics for Contextual Autonomous Driving Systems\\

\vspace{-0.3cm}

}

\author{\IEEEauthorblockN{Marie-Ngoïe Badibanga Kalenda}
\IEEEauthorblockA{\textit{Université de Technologie de Compiègne} \\
\textit{Ampere Software Technology} \\
mbadiban@utc.fr}
\and
\IEEEauthorblockN{Philippe Bonnifait}
\IEEEauthorblockA{\textit{Université de Technologie de Compiègne} \\
\textit{CNRS, Heudiasyc} \\
philippe.bonnifait@hds.utc.fr}
\and
\IEEEauthorblockN{Marie-Anne Mittet}
\IEEEauthorblockA{\textit{Ampere Software Technology} \\
Guyancourt, France \\
marie-anne.n.mittet@ampere.cars}

}

\maketitle

\begin{abstract}

Ensuring safety in autonomous driving requires continuous map maintenance supported by reliable quality indicators. In this context, it is crucial to identify when and where map updates should be triggered, for instance through crowdsourced data, and under which conditions a new map compilation should be deployed. This paper focuses on effective metrics for assessing the quality of vector maps and guiding such decisions. 
We present a new metric called $\text{GOSPA}_M$ designed to measure map discrepancies in terms of location errors, existence, and completeness. Through detailed simulations on both point and polyline feature maps, we analyze its sensitivity to common map degradation such as bias, false positives, false negatives, and coordinate errors. The results demonstrate that $\text{GOSPA}_M$ offers a unified and interpretable measure that effectively captures various forms of map deviation, making it a strong candidate for map quality assessment in automotive applications.

\end{abstract}


\section{Introduction}

\begin{figure}[h!]
    \centering
    \includegraphics[width=0.8\linewidth]{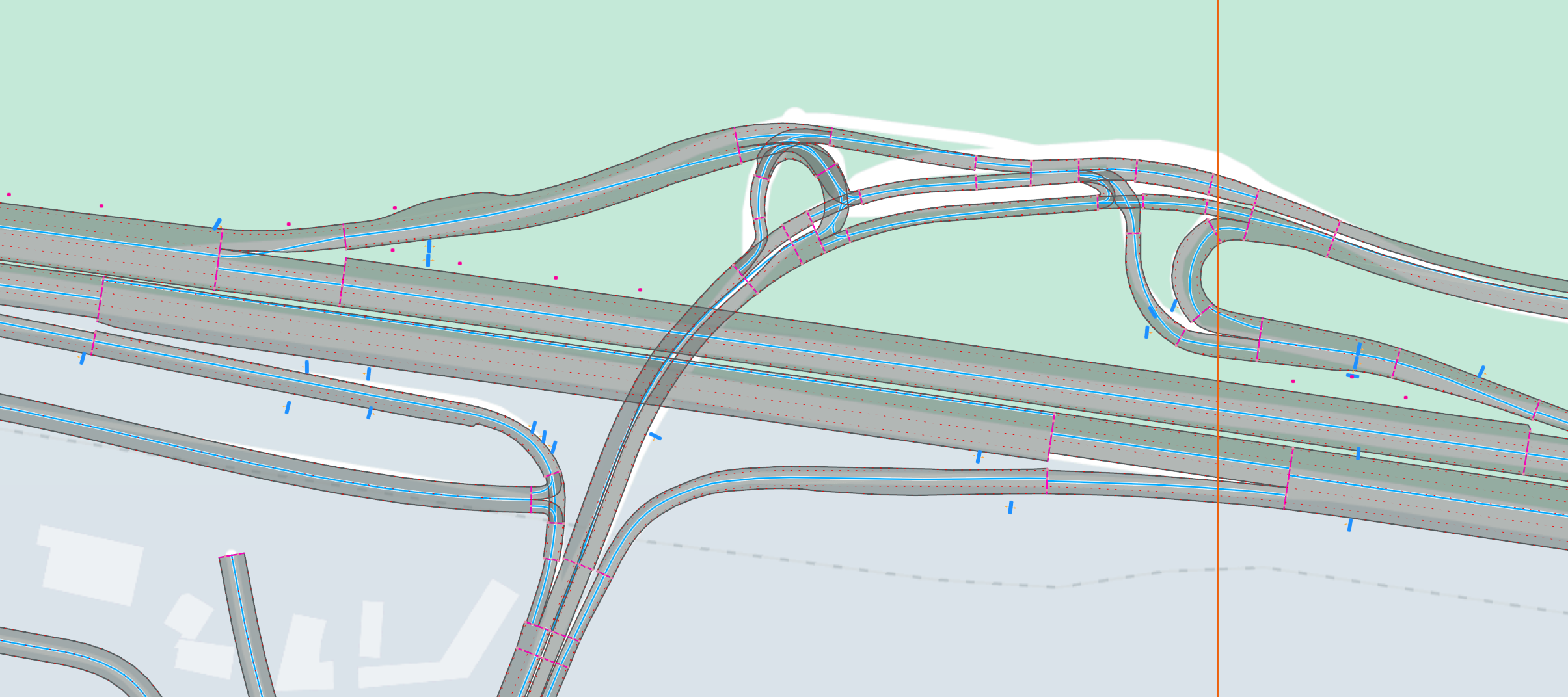}
    \caption{HD vector Map example (from Here)}
    \label{fig:Map example}
    \vspace{-0.6cm}
\end{figure}

The reliability and accuracy of vector maps (Fig.~\ref{fig:Map example}) are central for the development and deployment of autonomous driving systems. These systems rely on detailed digital representations of the road environment to take decisions in real time. However, the presence of errors, omissions or obsolete data in maps can compromise the safety, efficiency and robustness of autonomous vehicles.

In the rapidly evolving automotive industry, the maintenance of digital maps has undergone a significant transformation, largely influenced by the widespread adoption of Mobile Mapping Systems (MMS) \cite{Elhashash}. This shift has given rise to a new era of challenges marked by uncertainties and complexities, driven by the integration of crowdsourcing initiatives. The complexities arising from the variability in sensor performances, diverse vehicle fleets and varied environmental conditions \cite{Stephany_Berrio_Map_Maintenance} emphasize the need for a sophisticated approach to ensure accurary and effective mitigation strategies.

Recent surveys \cite{wijaya_high_2024, boubakri_high_2022} have highlighted that the lifecycle of HD maps—especially for autonomous driving systems—relies heavily on continuous updates to reflect real-world changes in road infrastructure and traffic signage data. Traditional map update approaches, which depend on high-precision MMS, are increasingly  replaced by crowdsourced strategies that leverage fleets of connected vehicles \cite{kim_hd_2021, dabeer_end--end_2017}. These vehicles, though equipped with lower-cost sensors, can collectively generate high-density data, enabling scalable and cost-effective maintenance \cite{pannen_how_2020}. 
Crowdsourcing use the vehicle sensors data of cars already on the roads.  Their sensors have lower performances, but the idea of crowdsourcing is that the high number of vehicles at different locations give the possibility to maintain the map at a good level.
However, challenges such as sensor variability, uncertain localization estimates, and partial feature visibility remain major obstacles to ensuring consistency, accuracy, and timeliness of map updates \cite{wong_mapping_2020}.

To address these challenges, multiple strategies have been proposed in recent literature. For example, feature-level patching of maps using graph-based SLAM and road model inference has been explored to correct outdated or missing structures \cite{pannen_how_2020}. Advanced filtering and matching algorithms, such as DBSCAN and KD-Tree \cite{moawad_offline_nodate}, have been employed to detect outliers and efficiently compare incoming observations with existing map data. More recently, mathematical models for systematic map updating have emerged, seeking to formalize when and where to apply updates \cite{guo_review_2024}. Despite these efforts, a unified metric capable of capturing geometric errors, missing detections, and semantic inconsistencies in a single framework remains largely absent. This motivates our exploration of new metrics, which aim to bridge this gap by offering a unified and interpretable tool to HD map evaluation and maintenance.

This research paper delves into the critical importance of employing a diverse range of metrics, both traditional and innovative, to monitor and address uncertainties in automotive map maintenance effectively.
We focus on identifying metrics that can be used to assess the quality of a given map. In particular, we present a new metric that is called $\text{GOSPA}_M$.

\section{Vector Maps}
On a 2D vector map, there are usually three types of objects (see Fig. \ref{fig:Map example}):
\begin{itemize}
    \item Punctual objects: e.g. traffic signs,
    \item Linear objects (links or polylines,): e.g. lane markings,
    \item Surface objects: e.g. buildings.
\end{itemize}

\begin{figure}[h!]
  \centering
  % Afficher la page 1 du fichier PDF
  \includegraphics[width=0.6\linewidth]{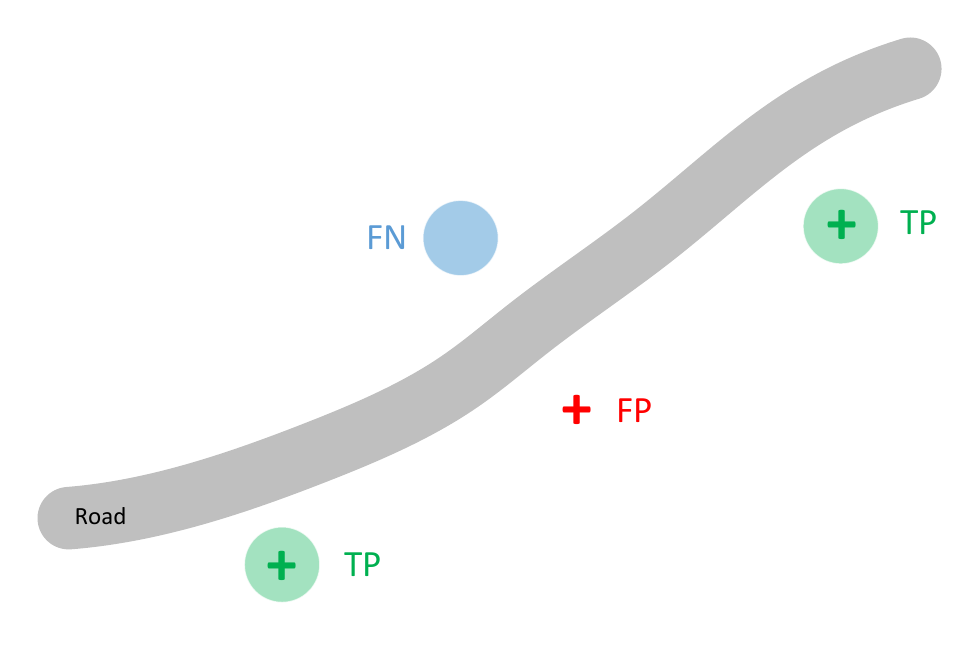}
  \caption{Illustration of the problem with a map of points. The crosses (+) are the map to be evaluated and the solid discs correspond to the real world.}
  \label{fig:Map description}
\end{figure}

Fig. \ref{fig:Map description} gives an example of what  happens during the maintenance of a map.
Each feature can be a:
\begin{itemize}
    \item True Positive (TP): the feature really exists and is in agreement with the real world (Green cross in  Fig. \ref{fig:Map description}) %good ones
    \item False Positive (FP): the feature does not exist in the real world (Red cross in  Fig. \ref{fig:Map description}) %Ghost
    \item False Negative (FN): a feature that exists in the real world that is not represented in the map (Blue disc in  Fig. \ref{fig:Map description}) %unmapped feature or missed
\end{itemize}

Since the free space is not explicitly represented in a vector map, True Negatives (TN) cannot be evaluated.

\begin{table}[ht]
    \centering
    \begin{tabular}{cccc}
        & Map feature & No Map feature \\
       Truly exists  & \cellcolor{customGreen} TP & \cellcolor{customBlue} FN \\
       Does not exist &\cellcolor{red!60} FP & \cellcolor{gray!50}TN \\
    \end{tabular}
    \caption{Map features evaluation logic}
    \label{tab:Basic table of evaluation metric}
\end{table}

When analyzing a map, three main aspects need to be evaluated :  feature existence,  feature location and  feature semantic.

Technically, a  good navigation map should have :
\begin{itemize}
    \item A high semantic exactness
    \item A high metric exactitude
    \item A low number of FNs
    \item A low number of FPs
\end{itemize}

All these elements  need to be measured and evaluated through metrics.

\section{Metrics}

\subsection{Methodology}

Let's consider that we have access to a reference map (also called Ground Truth) whose quality and accuracy are far better than the map to be evaluated. 
The objective is to identify relevant and interpretable metrics to qualify the quality of the latter.

Consider a set of true features from the reference map: \(X = \{x_1, x_2, ..., x_i\}\). The features from the map to be evaluated are provided in the following set: \(Y = \{y_1, y_2, ..., y_j\}\)

$x_i$ and $y_i$ are for example points in 2D coordinates. 

Please note that the two maps have in general different cardinalities (the number of elements contained in the sets). 

The first step in the evaluation process is to consider a limited geographical area (usually called a “vector tile”). The map tile to be evaluated must be completely enclosed within the reference map tile.
The elements of the two maps are then associated using map-matching algorithms that depend on the type of object to be assigned and the computational capacity of the processing machine. In this paper, we consider map-matching algorithms that use a maximum association distance, which is referred to as “gating”, as in object tracking algorithms.
%on fait le Munkres et ensuite le gating (=rejet des associations trop lointaine)
Thus, if the distance between the point evaluated and the reference point is below the gating threshold, the two points are considered as the same point.
Once the association is complete, metrics can be calculated.

The semantic exactness can be quantified by the percentage of correct semantic information of the features of the map to be evaluated that match the reference map. The task simply involves reviewing every map element and evaluating its semantic exactness with the help of the reference map. For example, if a speed limit sign on the map to be evaluated has a value of 50km/h and the actual value is 80km/h, this is a Boolean exactness of 0.  The average of these values corresponds to the map's semantic exactness.  
As it is quite straightforward to evaluate, we won't consider it further in the rest of this paper. We therefore focus on metrics for existence and location errors. 
These are affected by:
\begin{itemize}
    \item Coordinate errors on individual features: Errors different from each other in the the georeferencing of features (e.g. road signs and lane markings). This is sometimes referred to as “local imprecision”.
    \item Bias (or offset) of the map: Common-mode error in georeferencing of features. This is sometimes referred to as “global inaccuracy”.  
    \item Missing features: these are FNs.
    \item Non-existing mapped feature: these are FPs. 
\end{itemize}

\subsection{Usual metrics}

In the map to be evaluated, the matched points are labelled TP and the others FP. In the reference map, unmatched points are FN.
It is then possible to calculate the "Precision", often used to evaluate the correctness of a model prediction in machine learning \cite{kagde_automatic_2023}:

     \begin{equation}\label{Precision}
    \text{Precision}=\frac{TP}{TP + FP}
     \end{equation}
    
It represents the proportion of the map features that exist in the real world.
A map that has a high Precision is a map that has a low proportion of FP. It is typically a map that was well surveyed and that is up-to-date. Precision therefore is linked to what is called "freshness" sometimes when the world changes (e.g. some features are removed). 

Equivalently, "Recall" is defined as:
  \begin{equation}\label{Recall}
    \text{Recall} =\frac{TP}{TP + FN}
    \end{equation}
In machine learning, it quantifies the ability of the model to recognize an object that has been learned. For maps, it represents the proportion of the map features that exist in the real world.
A map that has a high recall has a low proportion of FN. It is typically a map that is complete (or almost complete). In other words, recall characterizes the completeness of a map. 
   
The F1-score is also often used to combine Precision and Recall  in a single number. It corresponds to the harmonic mean of Precision and Recall \cite{kagde_automatic_2023}:
\begin{equation}
F_{1} = 2 \cdot \frac{\text{Precision} \cdot \text{Recall}}{\text{Precision} + \text{Recall}} \in [0,1]
\label{F1}
\end{equation}

\begin{equation}
F_{1} = \frac{2TP}{2TP + FP + FN}
\end{equation}

A map that has a high F1-score has a good compromise between precision and recall. It is typically a map that is consistent with reality.

Consider now only the TPs, i.e. the set of the map-matched features. We call $\gamma$ this set which contains only  the matched couples ${x_k, y_p}$ from $X$ and $Y$.

The Mean Square Error  of the matched features ($\text{MSE}_M$), is defined as:
\begin{equation}\label{RMSE_M}
\text{MSE}_M=\frac{\sum_{k,p \in \gamma} (x_k, y_p)^2}{\mid \gamma \mid}   
\end{equation}
where $|\gamma|$ is the cardinality of the matching set $\gamma$.
The Root Mean Square Error is defined as $\text{RMSE}_M=\sqrt{\text{MSE}_M}$. 

This metric quantifies the average metric error (also called metric accuracy) of the features of the map to be evaluated but it doesn't handle the FN and FP values (i.e. Precision or Recall). 

These metrics are classically used to evaluate vector maps, but they do not provide a single numerical value for the overall quality of a map. 
In \cite{desjardins:hal-04803771}, a first attempt in this direction was proposed, and showed its relevance for pole maps. In the following section, we propose a new generic metric for evaluating the quality of vector maps.

\subsection{$\text{GOSPA}_M$: A new metric}

To evaluate the performance of trackers for automotive perception systems, the authors of \cite{rahmathullah_generalized_2017} have proposed a metric called "GOSPA" as Generalized Optimal SubPattern Assignment.
This metric has the property of combining into a single real number the localization errors of the detected objets, the number of FN and the number of FP.

The GOSPA between the sets $X$ and $Y$ is defined as : 

\begin{equation}\label{GOSPA1}
\begin{split}
\text{GOSPA} =\ & \Bigg[ \min_{\gamma \in \Gamma} \Bigg( \sum_{i,j \in \gamma} d(x_i, y_j)^{p} \\
& \qquad + \frac{c^p}{2}\big( |X| + |Y| - 2|\gamma| \big) \Bigg) \Bigg]^{1/p}
\end{split}
\end{equation}

with :
\begin{itemize}
    \item $\gamma$ is a possible assignment (matching between the two maps) and $\Gamma$ is the set of all the possible assignments,
    \item \(y_{j}\) is a feature of the map associated to a true value \(x_i\),
    \item $d(x_i, y_j)^{p}$ is the distance between matched features,
    \item $p$ is the power of the used norm. If $p$=2, one uses the Euclidean distance, 
    \item $c$ is the cutting threshold (the gating distance of the association here). Elements of $X$ and of $Y$ that are not associated because they are too far from each other receive the same association cost $\frac{c^p}{2}$,
    \item $\mid  ~ \mid$ denotes the cardinality of a set.

\end{itemize}

Let us see now how GOSPA can be adapted to the problem of estimating map quality.

First, please note that if the two maps $X$ and $Y$ have the same number of features and if every feature has been matched, then $| \gamma|=|X|=|Y|$ and the GOSPA becomes simply the sum of the localization errors between the features of the two maps. 

If not,  $|\gamma| = TP$, $|X|=TP + FN$ and $|Y|=TP + FP$.

The $1/p$ power can be calculated before minimization and the GOSPA can be expressed with the FN and FP values:

\begin{equation}\label{GOSPA3}
\text{GOSPA} = \min_{\gamma \in \Gamma} \left( \sum_{i,j \in \gamma} d(x_i, y_j)^{p} + \frac{c^p}{2} (FN + FP) \right)^{\frac{1}{p}}
\end{equation}

On can notice that in Eq.(~\ref{GOSPA3}), GOSPA is computed as an optimization problem for assignment, the cost function being the term between the large brackets.

By choosing the L2-norm and after minimization, we have:

\begin{equation}\label{GOSPA4}
\text{GOSPA} = \left( \sum_{k,p \in \gamma} d(x_k, y_p)^{2} + \frac{c^2}{2} (FN + FP) \right)^{\frac{1}{2}}
\end{equation}

As GOSPA depends on how many elements are present in the sets. Maps with a greater number of corresponding elements will have correspondingly larger GOSPA values.  This poses a problem, as a poor-quality map with few features may have a smaller GOSPA than a much larger and more accurate one.

To avoid this problem, one solution is to normalize GOSPA. There are several solutions:
\begin{enumerate}
    \item Use the cardinality of the reference map
    \item Use the cardinality of the map to be evaluated
    \item Use the cardinality of the number of features in common.
\end{enumerate}

From our point of view, the third solution is the most appropriate, since a map with few features in common with the reference map is a map with low similarity and therefore a high GOSPA. 

So, we propose a new metric called $\text{GOSPA}_M$ (GOSPA for Maps) as follows:
\begin{equation}\label{GOSPA_M}
\text{GOSPA}_M  = \frac{\text{GOSPA}}{\sqrt{| \gamma |}} \text{ and } | \gamma |\neq 0
\end{equation}
An interesting point of this metric is that it is coherent with $\text{MSE}_M$ since it has the same normalization term.

Its expression after minimization becomes very simple with the L2-norm:
\begin{equation}\label{GOSPA_M2}
\text{GOSPA}_M = \sqrt{\text{MSE}_M +  c^2 \left( \frac{FN + FP}{2TP}\right)}  \text{ and } TP\neq 0
\end{equation}

Note that if the map contains no FN or FP, then the $\text{GOSPA}_M$ becomes the $\text{RMSE}_M$. The ratio $\frac{FN + FP}{2TP}$ can be interpreted as a feature existence and no-existence error rate. Please note that $\text{GOSPA}_M$ has a singularity when the map is fully wrong, i.e when there is no TP. If the tile is large enough, this singularity does not occur in practice.

$\text{GOSPA}_M$ is a measure of similarity between maps. It is expressed in meters. The lower its value, the more coherent the maps are.

\section{Case study}
In this section, we focus on two types of features contained in vector maps: points and polylines.

\subsection{Point features}
Points on the map generally correspond to road signs, traffic lights or georeferenced posts.

There are different possibilities in order to do the map-matching between two sets of points.
The first step is to choose a distance that can be the Euclidean distance or the Mahalanobis distance to compute associations costs. 
Euclidean distance is computationally more efficient and does not need the knowledge of covariance matrices. This is the distance that we have used in the results of this paper. 

The Munkres algorithm \cite{munkres1957algorithms}, also known as the Hungarian algorithm, is often used in optimization problems that require optimal one-to-one assignments—such as matching two sets of points in a way that minimizes an overall cost. When applied to point-to-point map matching, each potential match is assigned a cost. The Munkres algorithm then finds the assignment that minimizes the total cost and still maximizing the number of pairs \cite{xuemei_real-time_2024}.

Then, we consider that the map-matching is accepted if the chosen distance between the points of the map and the real ones is below a specific threshold (denoted $c$ in Eq.~(\ref{GOSPA_M2}) as done in \cite{dehelean_ospa_2018}.

To calculate $\text{GOSPA}_M$, it is then just a matter of computing $\text{RMSE}_M$, TP, FN and FP and applying Eq.~(\ref{GOSPA_M2}).

\subsection{Polyline features}

Another fundamental representations within vector maps is the use of polylines to model lanes. Each lane in a road network can be captured as a set of polylines (e.g. lane marking, center lane or road border). This level of granularity is especially important for modern automotive applications such as autonomous driving, advanced driver-assistance systems (ADAS), and detailed navigation tools, where understanding lane geometry and connectivity is critical.

Line-to-line matching is a technique often used to align sequences of spatial data (such as vehicle GPS traces) with corresponding line segments on a digital map, such as road or lane polylines. Unlike point-based map matching, which aligns individual positions to nearby features, line-to-line matching evaluates entire trajectory segments against mapped lane geometries. This provides a context-aware alignment, especially in complex urban environments where data may be noisy, ambiguous, or sparse. By assessing how well two line geometries match in both shape and direction, line-to-line matching allows lane-level localization. We use it here to evaluate the quality of the polylines of a map.

To perform line-to-line matching between two maps, a cost matrix is computed between all pairs of polylines. The metric that we use to populate this matrix is the weighted Frenet distance. We considered also the use of the Hausdorff distance that considers the worst-case deviation but this distance provides assignments that are less smooth and less shape-consistent.

The weighted Frenet distance given in Eq.~(\ref{weighted frenet distance}) measures the average lateral deviation between the points of a given line and their Frenets's projections on the  reference lines. Weighting is then added. Weights may be uniform or context-dependent, based on features such as curvature, confidence, or spatial sampling. This metric emphasizes the overall alignment and shape consistency of the polylines, making it less sensitive to isolated outliers.

\begin{equation}\label{weighted frenet distance}
    \centering
    \bar{d} = \frac{\sum_{i=1}^{n} w_i d_i}{\sum_{i=1}^{n} w_i}
\end{equation}
where $d_i$ is the cross-track distance in the Frenet's frame and $n$ the number of points of the polyline under evaluation.

Once the cost matrix is populated using the distance metric between the polylines of the two maps, the Munkres algorithm is again applied to determine the optimal assignment between the two maps. $\text{GOSPA}_M$ is computed using the matched polylines and the others in a similar way to points.

\section{Results}

We report here simulation results.
We took a tile from an automotive supplier's HD map. This map is the reference map. We have used QGIS (Quantum Geographic Information System) to manipulate it. QGIS is widely used in the automotive industry for visualizing and managing geospatial data, particularly for mapping road infrastructure. 
We degraded the reference map with some random metric errors, deleted elements and added elements to create a map to be evaluated. 
This enabled us to carry out a sensitivity study of $\text{GOSPA}_M$.
Our implementation was made in Python using the Gospapy library\footnote{\url{https://github.com/ewilthil/gospapy/tree/master}}.

For ease of analysis, we consider point maps and polyline maps separately.
For each case, we consider only one parameter at a time (i.e. coordinate errors, bias, FN and FP), so that the $\text{GOSPA}_M$ sensitivity can be analyzed individually. The gating distance in Eq.~(\ref{GOSPA_M2}) was set at $c=2$~m in all the experiments, as it corresponds to the global accuracy of the map.

\subsection{Point features map}

\begin{figure}
    \centering
    \includegraphics[angle=0,width=1\linewidth]{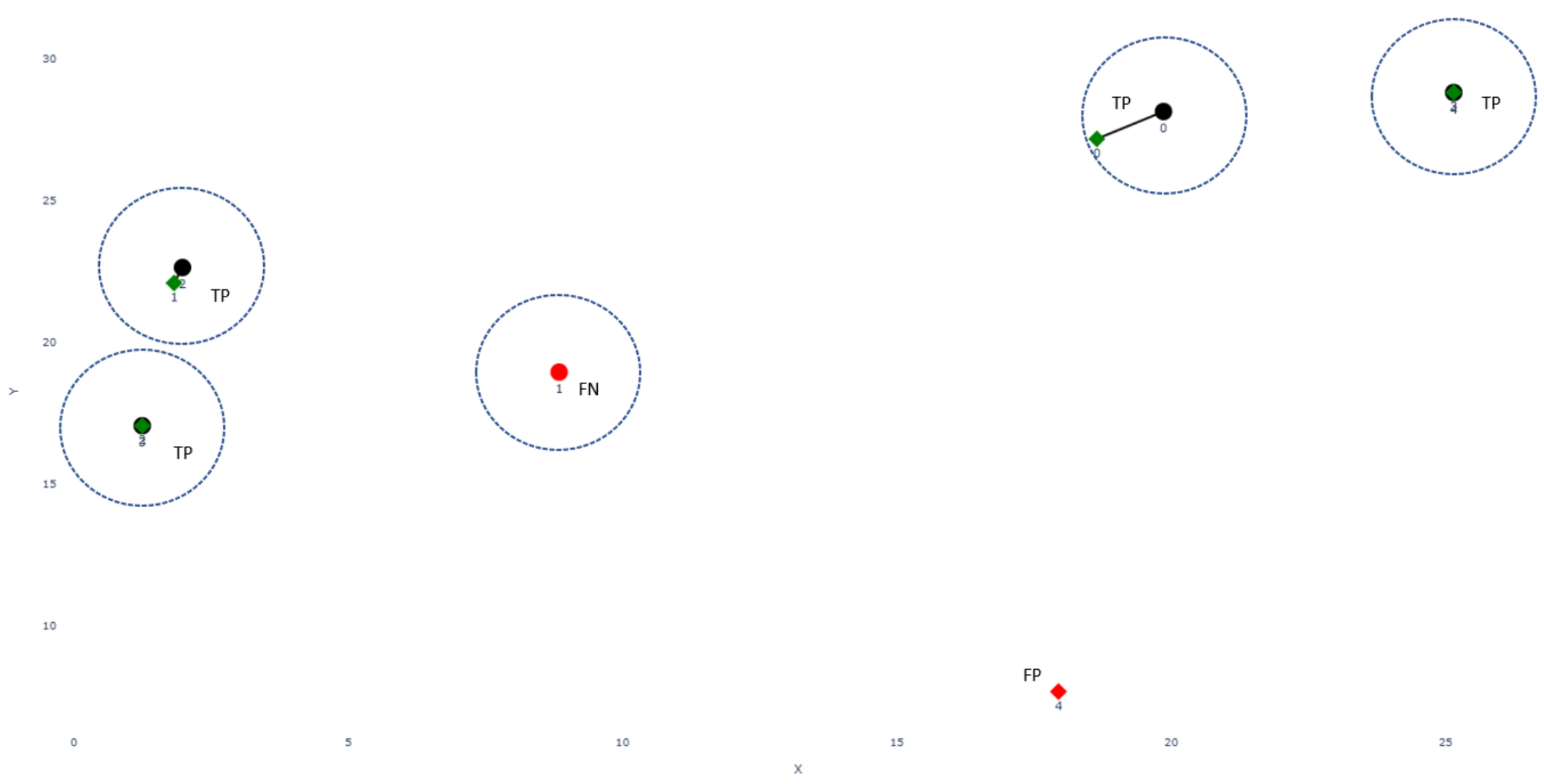}
    \caption{Example of tile with point features. The circles correspond to the reference map, the diamonds to a map to be evaluated, the dotted circles show association gating and the black lines show associations. Units are meters.}
    \label{fig:Reference points}
\end{figure}

We have chosen to consider a reduced tile size of 30mx30m to have few features and simplify the analysis of the results (see Fig. \ref{fig:Reference points}). 

\begin{figure*}[htbp]
    \centering

    % Ligne 1
    \begin{subfigure}[t]{0.48\textwidth}
        \centering
        \includegraphics[width=\linewidth]{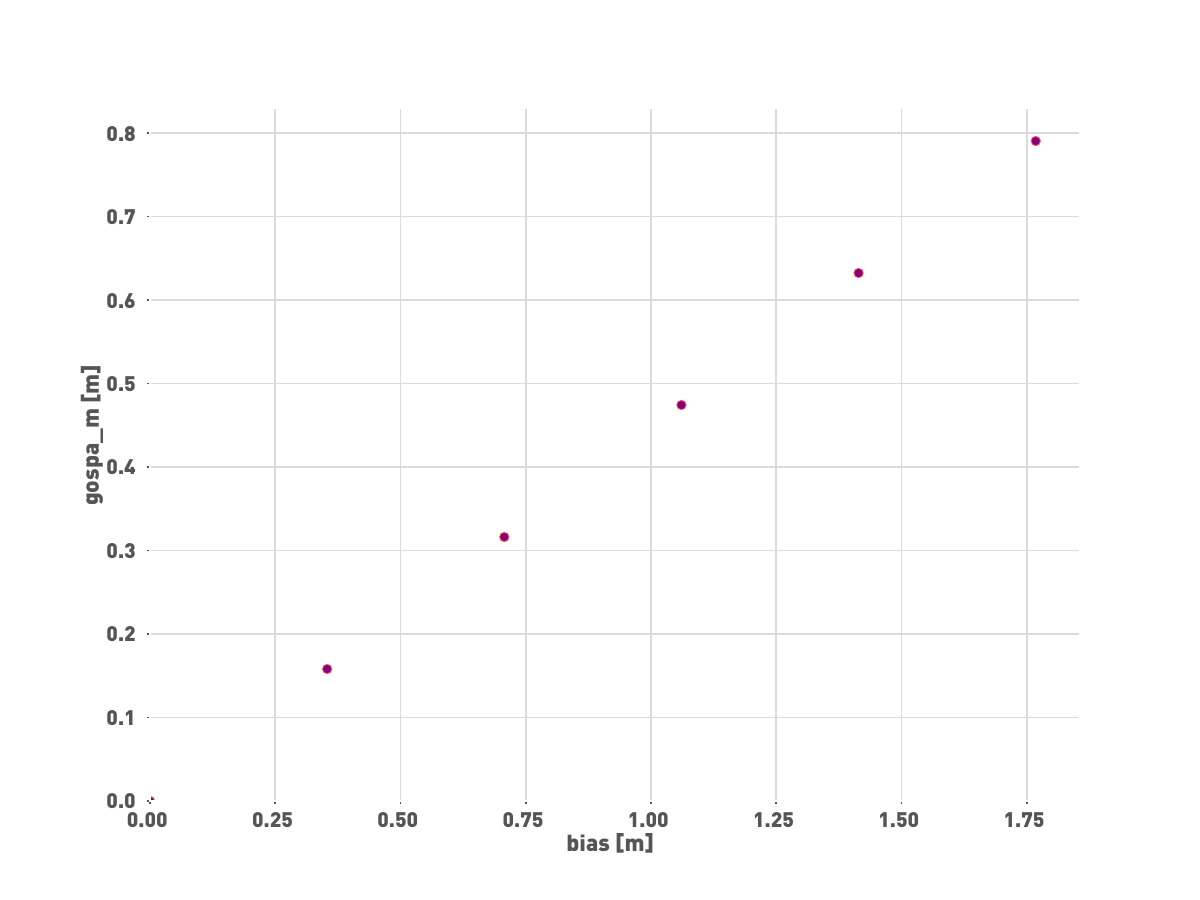}
        \label{fig:distance_vs_gospa}
    \end{subfigure}%
    \hfill
    \begin{subfigure}[t]{0.48\textwidth}
        \centering
        \includegraphics[width=\linewidth]{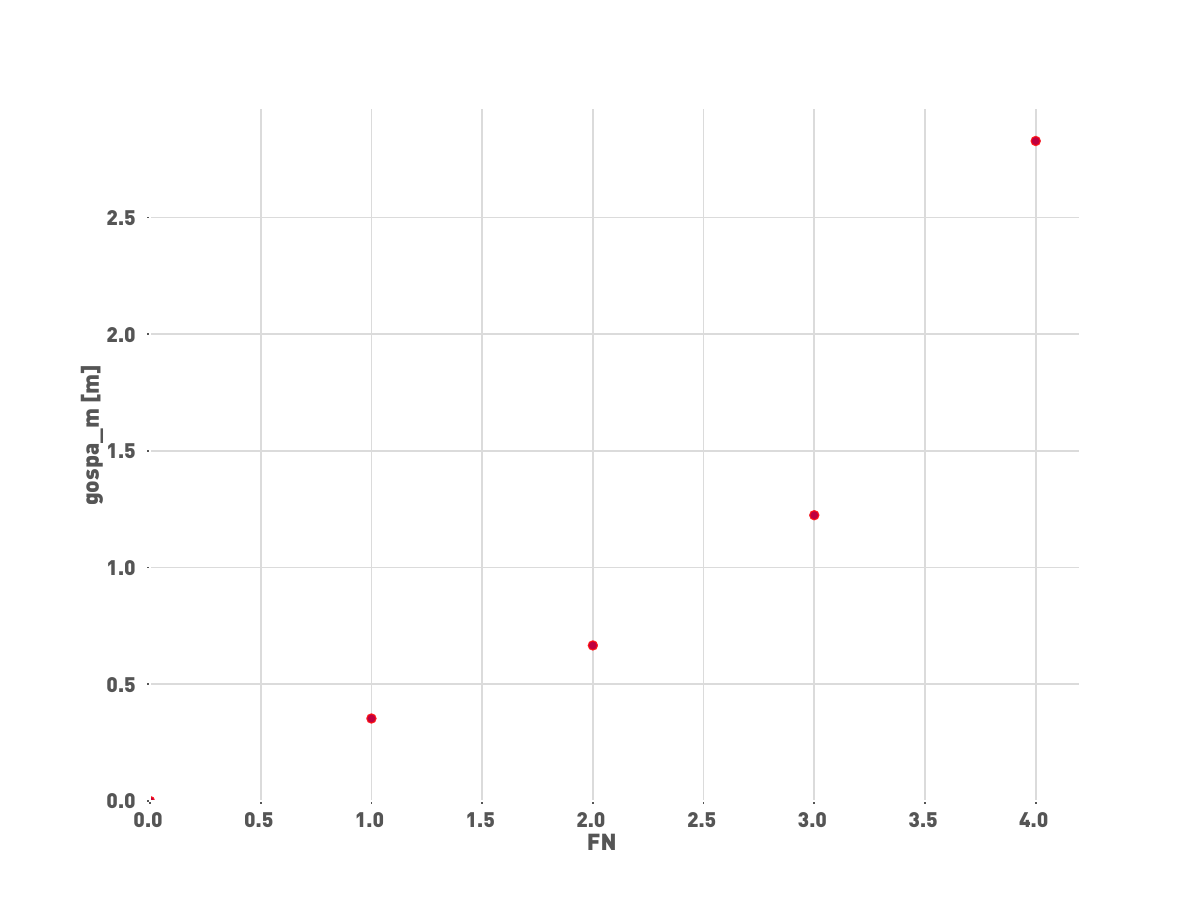}
        \label{fig:n_FN_vs_gospa}
    \end{subfigure}\\[1ex]
    % Ligne 2
    \begin{subfigure}[t]{0.48\textwidth}
        \centering
        \includegraphics[width=\linewidth]{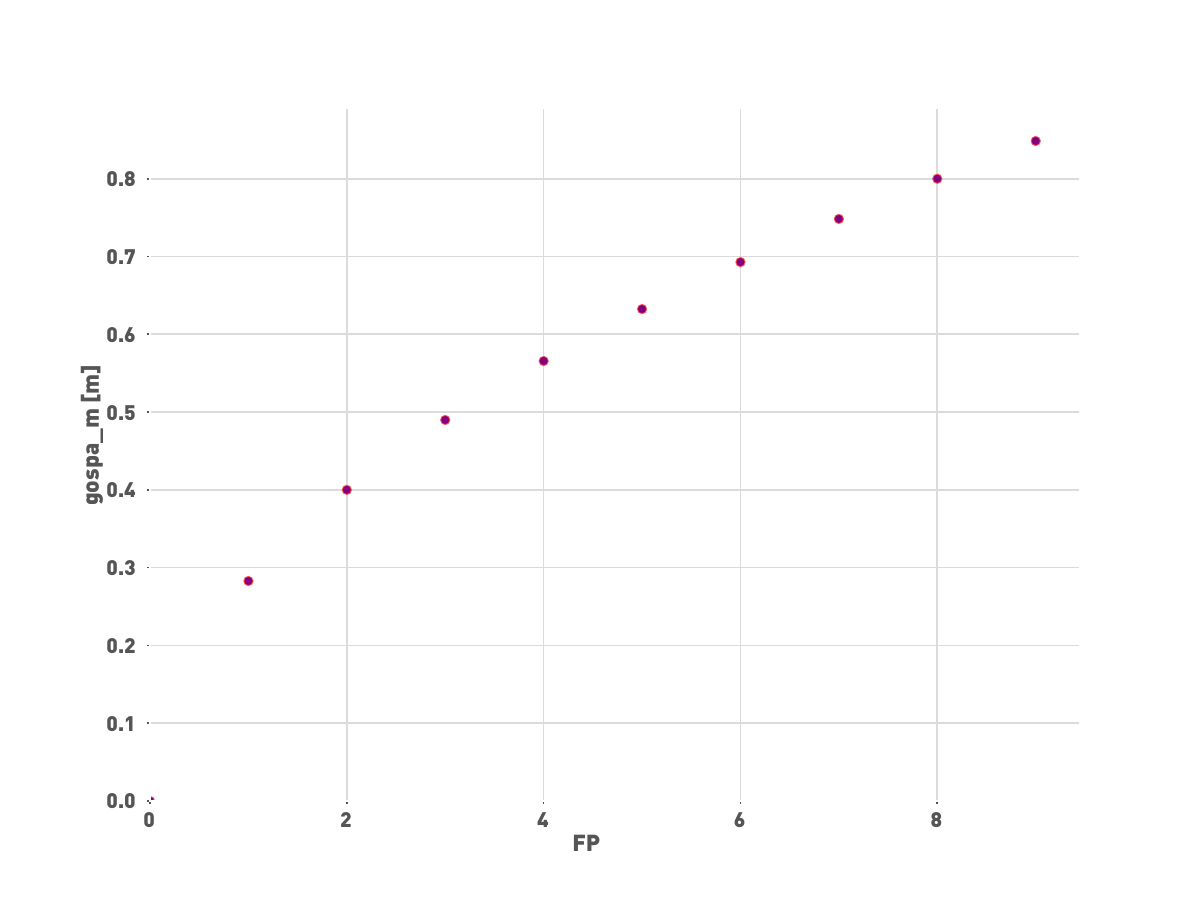}
        \label{fig:nb_FP_l_vs_gospa}
    \end{subfigure}%
    \hfill
    \begin{subfigure}[t]{0.48\textwidth}
        \centering
        \includegraphics[width=\linewidth]{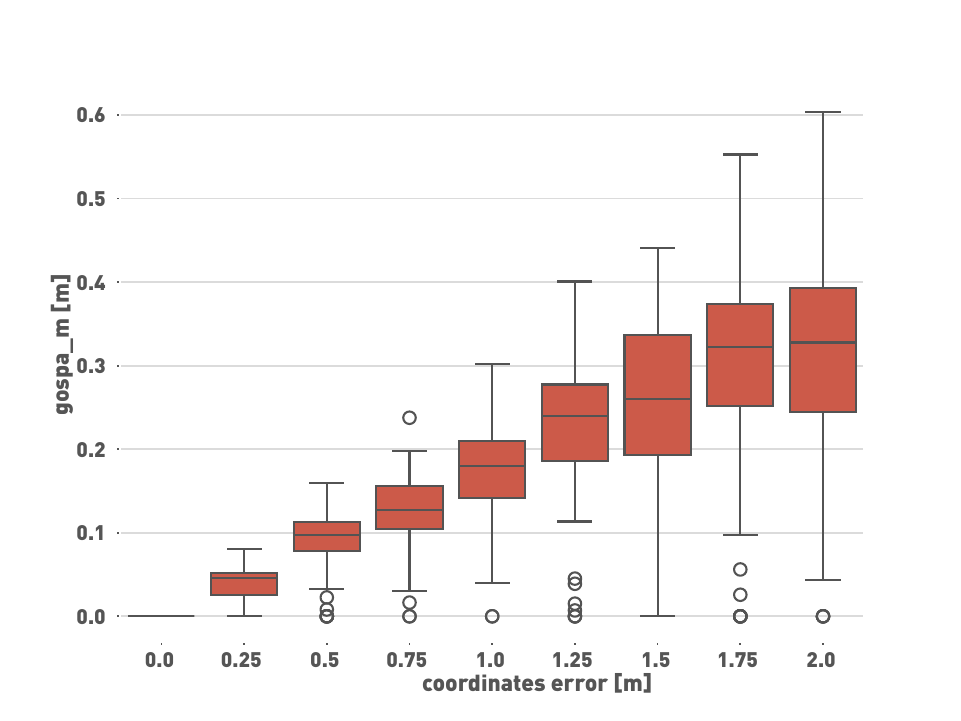}
        \label{fig:box_plot_gospa}
    \end{subfigure}
      
    \caption{From top to bottom  and left to right: effects of the bias, FN, FP, and coordinate errors (Monte Carlo box plot) on \(Gospa_M\) for for a map containing only points }
    \label{fig:global_analysis for Points-only map}
\end{figure*}

\subsubsection{Bias variation}

This test was done by varying the value of bias from 0 to 2 meters in order to keep it below the gating distance. In the test, we used the same distance increment value for \(d_x\) and \(d_y\). The resulting bias is \( \sqrt{d_x^2 + d_y^2}\).

The first subplot of Fig.~\ref{fig:global_analysis for Points-only map} displays the evolution of the $\text{GOSPA}_M$. Its value keeps increasing as the bias increases.
Note that when the bias is higher than 2~m, there is no TP, which means that the $\text{GOSPA}_M$ cannot be calculated. This explains why we kept its value under 1.75~m. 
As there is no $FN$ and no $FP$, $\text{GOSPA}_M$ is identical to $\text{RMSE}_M$. In this simulation, the bias evolves linearly and identically. Indeed, the bias comes out of the square root of Eq.~(\ref{GOSPA_M2}) as it is positive and $\text{GOSPA}_M$ evolves linearly with the bias as observed.

\subsubsection{FN variation}

In this test, there is no $FP$ and no $\text{MSE}_M$.  When $FN=1$, the value of $TP$ is reduced by 1. 
So, considering $TP_{max}$, the maximum number of TPs with a perfect map, Eq.~(\ref{GOSPA_M2}) becomes:

\begin{equation}\label{TPmax}
\text{GOSPA}_M = c \sqrt{\frac{FN}{2(TP_{max}-FN)}}=f(FN) 
\end{equation}
In order to calculate its value there is a necessity to still have at least one $TP$ value. That's why $FN$ is limited to 4 here. 
Eq.~\ref{TPmax} explains the non-linear shape of the corresponding subplot in Fig.~\ref{fig:global_analysis for Points-only map}. 

\subsubsection{FP variation}

When the $FP$ varies, ghosts points are added. $TP$ remains constant and there is no FN and no $\text{MSE}_M$. There is no technical limits to the number of $FP$ applicable. $\text{GOSPA}_M \propto \sqrt{FP}$
as shown by the subplot of Fig.~\ref{fig:global_analysis for Points-only map}.

\subsubsection{Coordinates error variation}

To simulate location  errors,  random noises were added to the coordinates of the points. Their values were generated between zero and the maximum error corresponding to the gating distance of 2 meters to avoid the generation of FP and FN. Each point was modified by an error different from the others. 
A single simulation being not relevant, we carried out a Monte Carlo study (100 tests for each value).
In Fig.~\ref{fig:global_analysis for Points-only map}, box plots are displayed. The median shows a non-linear trend such as a square root function. % $\propto \sqrt{e}$. 
We also observe that the greater the errors, the wider the dispersion of $\text{GOSPA}_M$ around the medians. These curves are consistent with the evolution of RMSE as a function of a random parameter.
\subsection{Polyline features map}

\begin{figure}[H]
    \centering

        \includegraphics[width=0.999\linewidth]{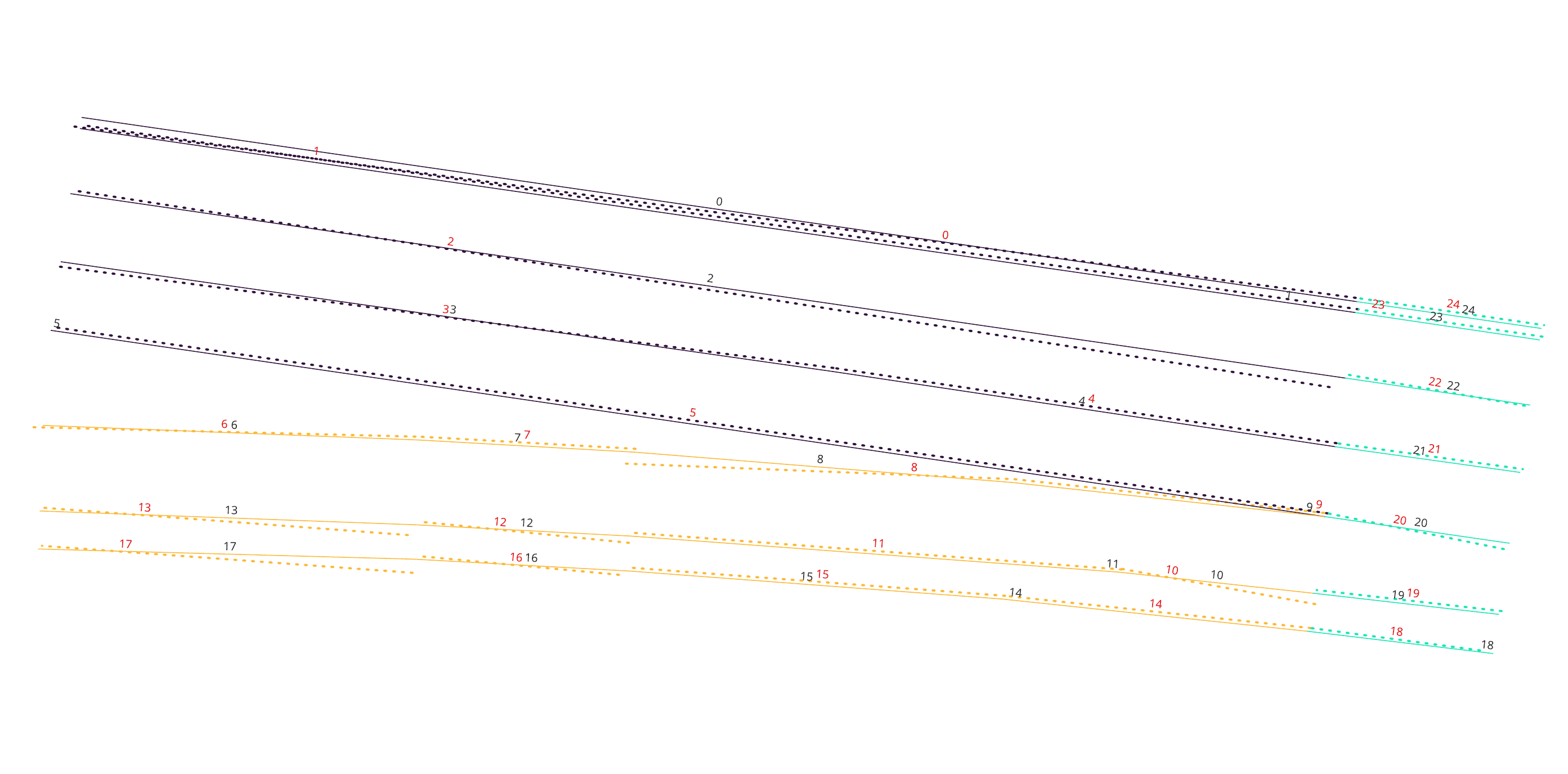}

    \caption{Illustration of the two polyline maps. The map to be evaluated is in dotted lines. The colors show the existence of 3 lanes groups here.}
    \label{fig:map_dx_comparison}
\end{figure}

In this study, the map in dotted lines  was analyzed with respect to the reference map in continuous lines (see Fig. \ref{fig:map_dx_comparison}).

\begin{figure}[htbp]
    \centering

    % ---------- Row 1 ----------

        \includegraphics[width=1\linewidth]{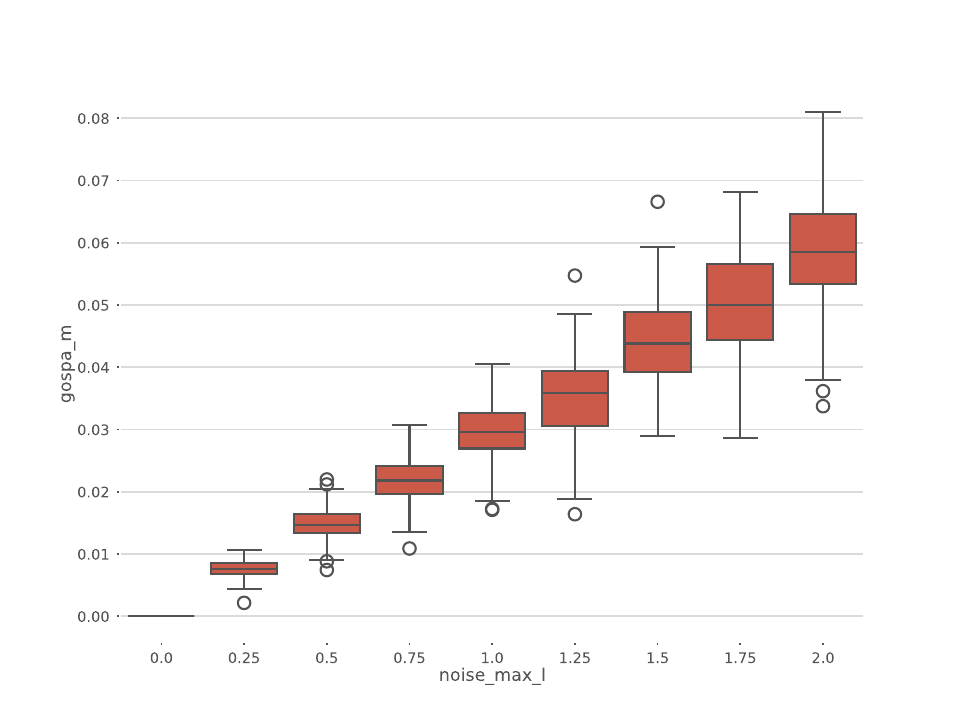}
        \caption{$\text{GOSPA}_M$ as a function of the coordinate errors for a map containing only polylines.}
        \label{fig:noise_max_Gospa_M}

\end{figure}

\subsubsection{Bias variation}
During the tests, the bias evolved from 0 to 2 meters. We have noticed that the results are fully similar to the point features case of Fig. \ref{fig:global_analysis for Points-only map}. $\text{GOSPA}_M$ has a linear evolution as a function of bias.

\subsubsection{FN variation}

In this case, the FNs are randomly applied in the map, no matter the group lane or the lines around.
The results are entirely similar to the point features map (Fig. \ref{fig:global_analysis for Points-only map}).

\subsubsection{FP variation}
The evolution of \(Gospa_M\) when the number of FPs changes is again the same as in Fig. \ref{fig:global_analysis for Points-only map}. 

\subsubsection{Coordinates error variation}
In this case, there is a little change between the point features map and the polyline features map. Please see the box plots of Fig. \ref{fig:noise_max_Gospa_M}.
Even if the median trend remains the same, the $\text{GOSPA}_M$ for polylines is less spread in terms of box size. It is explained by the fact that for polylines more points have to be affected by errors to get the same random variation. It means that the computation of $\text{GOSPA}_M$ for polylines has a kind of smoothing property.

\subsection{Discussion}

Globally, $\text{GOSPA}_M$ shows a linear response to bias, an explainable evolution with FN and FP variations, and a dispersion behavior under coordinate noise consistent with RMSE-like metrics. Additionally, in the case of polyline features, $\text{GOSPA}_M$ demonstrated a smoothing effect, reinforcing its robustness in more complex map structures.
It can be noticed that $\text{GOSPA}_M$ a sensitivity of the same order of magnitude in all directions. This property is very nice because, in practice, the various sources of error are combined together. In other words, the mathematical expression of $\text{GOSPA}_M$ is well balanced between the different input parameters, so that none of them will have a dominant effect over the others.

Today, some authors such as \cite{bock_unsettled_2021} have observed that 80\% of a road map remains stable and that only 7 to 15\% of it undergoes changes. Even if a map always contains errors, the goal is to keep it sufficiently representative of the environment for the ADAS functions that use it \cite{yang_review_2024}. 
The value $\text{GOSPA}_M$ can be 0 when the map is perfect, but its value is not bounded. Further studies to complement the results presented in this paper are needed to identify which $\text{GOSPA}_M$ values correspond to a map of sufficient quality.

\section{Conclusion and Perspectives}

This article presented several metrics for assessing the quality of vector automotive maps containing points and/or polylines. In particular, we proposed a new metric, $\text{GOSPA}_M$, which synthesizes key aspects of map quality into a single real value.
Tests conducted on both point and polyline features demonstrated the relevance of this metric for capturing geospatial discrepancies. Through controlled simulations, we showed that $\text{GOSPA}_M$ is sensitive to different types of perturbations—systematic (e.g., bias), stochastic (e.g., random noise), and structural (e.g., false positives/negatives)—making it well-suited for map evaluation tasks.
Its behavior remains interpretable across various degradations and well-balanced with respect to its inputs, reinforcing its usefulness for monitoring HD map quality in the automotive domain.

Maintaining up-to-date and accurate maps is  crucial for vehicle safety. High-quality maps help perception and planning systems interpret the driving environment more effectively, anticipate road features, and make safer decisions. Thus, a low $\text{GOSPA}_M$ score reflects not only better map fidelity but also contributes directly to safer autonomous and assisted driving.

Future work includes using $\text{GOSPA}_M$ as a reference metric to evaluate and compare various crowdsourced map updating strategies. Applying it to real-world maps will also allow validation under real-life complexity and update requirements.
Finally, even though $\text{GOSPA}_M$ is essentially a relative metric for comparing maps with each other or evaluating degradation or improvement over time, we plan to conduct case studies to identify ranges of values that could be used to qualify the overall quality of a map in absolute terms.

\printbibliography

\end{document}